\documentclass[lettersize,journal]{IEEEtran}
\usepackage{amsmath,amsfonts}
\usepackage{algorithmic}
\usepackage{algorithm}
\usepackage{array}
\usepackage[caption=false,font=normalsize,labelfont=sf,textfont=sf]{subfig}
\usepackage{textcomp}
\usepackage{stfloats}
\usepackage{url}
\usepackage{verbatim}
\usepackage{graphicx}
\usepackage{amsmath}
\usepackage{makecell}
\usepackage{xcolor}
\usepackage{booktabs} 
\usepackage{multirow} 
\definecolor{mustard}{HTML}{FFDB58}

\begin{document}

\title{BREATH-VL: Vision-Language-Guided 6-DoF Bronchoscopy Localization via Semantic-Geometric Fusion}


\author{Qingyao Tian, Bingyu Yang, Huai Liao, Xinyan Huang, Junyong Li, Dong Yi and Hongbin Liu
\thanks{Qingyao Tian and Bingyu Yang are with State Key Laboratory of Multimodal Artificial Intelligence Systems, Institute of Automation, Chinese Academy of Sciences, Beijing 100190, China, and also with the School of Artificial Intelligence, University of Chinese Academy of Sciences, Beijing 100049, China.}%
\thanks{Huai Liao, M.D. and Xinyan Huang, M.D. are with Department of Pulmonary and Critical Care Medicine, The First Affiliated Hospital of Sun Yat-sen University, Guangzhou, Guangdong Province, P.R. China.}%
\thanks{Junyong Li and Dong Yi are with Centre of AI and Robotics, Hong Kong Institute of Science \& Innovation, Chinese Academy of Sciences.}%
\thanks{Corresponding author: Hongbin Liu is with Institute of Automation, Chinese Academy of Sciences, and with Centre of AI and Robotics, Hong Kong Institute of Science \& Innovation, Chinese Academy of Sciences. He is also affiliated with the School of Biomedical Engineering and Imaging Sciences, King’s College London, UK. (e-mail: liuhongbin@ia.ac.cn).}%
}

\markboth{Journal of \LaTeX\ Class Files,~Vol.~14, No.~8, August~2021}%
{Shell \MakeLowercase{\textit{et al.}}: A Sample Article Using IEEEtran.cls for IEEE Journals}


\maketitle
\begin{abstract}

Vision-language models (VLMs) have recently shown remarkable performance in navigation and localization tasks by leveraging large-scale pretraining for semantic understanding. However, applying VLMs to 6-DoF endoscopic camera localization presents several challenges: 1) the lack of large-scale, high-quality, densely annotated, and localization-oriented vision-language datasets in real-world medical settings; 2) limited capability for fine-grained pose regression; and 3) high computational latency when extracting temporal features from past frames. To address these issues, we first construct BREATH dataset, the largest in-vivo endoscopic localization dataset to date, collected in the complex human airway. Building on this dataset, we propose \textbf{BREATH-VL}, a hybrid framework that integrates semantic cues from VLMs with geometric information from vision-based registration methods for accurate 6-DoF pose estimation. Our motivation lies in the complementary strengths of both approaches: VLMs offer generalizable semantic understanding, while registration methods provide precise geometric alignment. To further enhance the VLM’s ability to capture temporal context, we introduce a lightweight context-learning mechanism that encodes motion history as linguistic prompts, enabling efficient temporal reasoning without expensive video-level computation. Extensive experiments demonstrate that the vision-language module delivers robust semantic localization in challenging surgical scenes. Building on this, our BREATH-VL outperforms state-of-the-art vision-only localization methods in both accuracy and generalization, reducing translational error by 25.5\% compared with the best-performing baseline, while achieving competitive computational latency.
\end{abstract}

    
\section{Introduction}
\label{sec:intro}
\begin{figure}[tp]
\centerline{\includegraphics[width=\columnwidth]{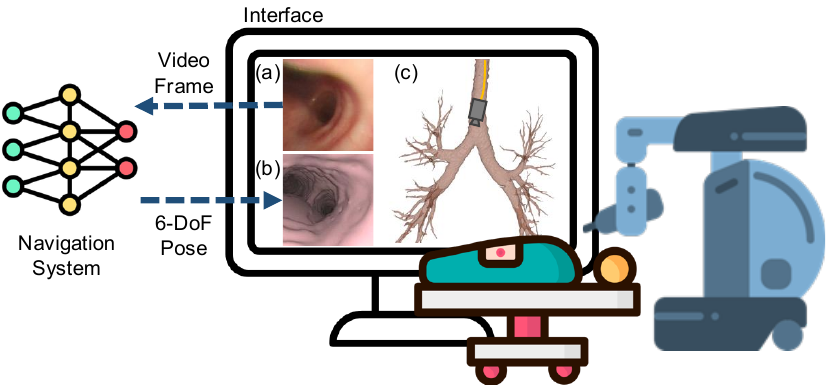}}
\caption{Clinical workflow of visually-assisted bronchoscopy navigation. During robotic or conventional bronchoscopy, the navigation system receives endoscopic video frames and estimates the 6-DoF pose of the endoscope, which is then used to provide visual feedback to the surgeon. (a) Bronchoscopic frame. (b) Virtual bronchoscopy view rendered at the estimated pose. (c) Global airway view showing the endoscope’s position within the patient-specific airway.
}

\label{fig:system}
\end{figure}

\begin{figure*}[tp]
\centerline{\includegraphics[width=\textwidth]{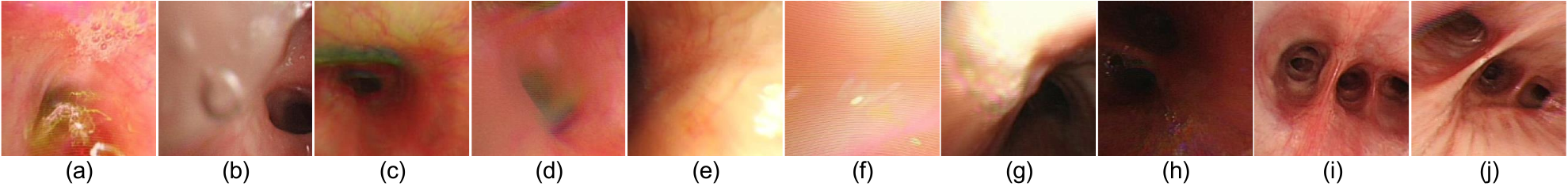}}
\caption{Challenging frames from the BREATH dataset. (a)-(b) show visual artifacts such as fluids and bubbles occluding the field of view. (c)-(d) show motion blur caused by rapid bronchoscope motion. (e)-(f) show textureless regions. (g)-(h) show illumination disturbances including high contrast and darkness. (i)-(j) show anatomically distinct airway regions with similar visual appearance, which can confuse landmark-based methods; (i) is from the left inferior lobar bronchus, and (j) is from the right intermediate bronchus.
}

\label{fig:dataset}
\end{figure*}

\IEEEPARstart{V}{isually}-navigated interventional surgery can provide accurate, low-cost guidance to surgeons with minimal setup. Figure~\ref{fig:system} illustrates the clinical workflow of visually-navigated bronchoscopy. In these settings, prior work has primarily focused on vision-only methods for surgical localization and navigation \cite{ozyoruk2021endoslam,banach2021visually,sganga2019autonomous,banach2025conditional}. However, endoscopic localization poses unique challenges: images are often degraded by fluid occlusions and motion blur; contain textureless or feature-poor regions; illumination is complex; and anatomical structures are highly deformable and repetitive. Figure~\ref{fig:dataset} shows bronchoscopic examples illustrating these challenges. These conditions make vision-based localization extremely difficult, highlighting the need for intelligent, context-aware methods that can reason about anatomy and motion beyond purely geometric cues.

Vision-language models (VLMs) have recently gained attention for localization \cite{xu2024addressclip,jia2025towards,cheng2025scale,xu2025addressvlm,li2024georeasoner,zhang2024can} and navigation \cite{an2024etpnav,tian2025drivevlm,yokoyama2025film,chen2021topological} tasks due to their ability to integrate high-level semantic understanding into visual perception. By aligning visual inputs with language, VLMs can provide contextual priors \cite{chen2021topological}, reduce visual ambiguity \cite{xiao2025mvl}, support zero-shot generalization to unseen environments \cite{zhang2024can}, and guide estimation using language-based instructions \cite{li2024georeasoner}. These capabilities offer a strong complement to vision-based methods, leading to more robust and generalizable pose estimation in complex or ambiguous scenes.

Despite advances in natural environments, the potential of VLMs in interventional and surgical domains remains largely unexplored. Motivated by the success of VLMs in natural-scene localization \cite{xu2025addressvlm,xu2024addressclip}, we study their use for assisting 6-DoF bronchoscopy camera localization. However, deploying VLMs in surgery raises three challenges: 1) unlike natural scenes, bronchoscopy lacks large-scale, domain-specific training data, which limits semantic understanding; 2) VLMs are not designed for fine-grained pose regression, making them unsuitable as drop-in replacements for existing navigation workflows; and 3) temporal cues are crucial for surgical navigation \cite{tian2025endomamba}, but providing video clips to VLMs is computationally heavy and impractical in surgical pipelines.

\begin{table*}[t]
\centering
\caption{Comparison of publicly available surgical endoscopic datasets for localization, reconstruction, and visual odometry.}
\label{tab:datasets}
\renewcommand{\arraystretch}{1.2}
\setlength{\tabcolsep}{4pt}
\footnotesize
\begin{tabular}{lccccc}
\Xhline{1.2pt}
\textbf{Dataset} & \textbf{Organ / Region} & \textbf{Videos / Sequences} & \textbf{Labeled Frames} & \textbf{Type} & \textbf{Purpose} \\
\Xhline{1.0pt}

{EndoMapper} \cite{azagra2023endomapper}
& Colon & 96 & 286{,}707 & In-vivo & VSLAM \\
\cline{3-5}
&  & 5 & 1{,}992 & Simulation &  \\
\Xhline{0.9pt}

{C3VD} \cite{bobrow2023}
& Colon & 26 & 37.8k & Simulation & Reconstruction \\
\Xhline{0.9pt}

{C3VDv2} \cite{golhar2025c3vdv2}
& Colon & 8 & 95{,}300 & Simulation & Reconstruction \\
\cline{3-5}
&  & 192 & 169{,}371 & Phantom &  \\
\Xhline{0.9pt}

{EndoSLAM} \cite{ozyoruk2021endoslam}
& Colon, Stomach, & 58 & 42{,}700 & Ex-vivo & Reconstruction \\
\cline{3-5}
& Small Intestine & 3 & 35{,}993 & Simulation &  \\
\Xhline{0.9pt}

{SimCOL3D} \cite{rau2024simcol3d}
& Colon & 33 & 23{,}421 & Simulation & Depth and Pose Estimation \\
\cline{3-6}
&  & 59 & -- & In-vivo & Pose Estimation \\
\Xhline{0.9pt}

{Fulton et al.} \cite{fulton2020comparing}
& Colon & 7 & -- & Simulation & Visual Odometry \\
\Xhline{0.9pt}

{Deng et al.} \cite{deng2023feature}
& Airway & 27 & 17{,}398 & Phantom & Visual Odometry \\
\cline{3-6}
&  & -- & -- & Ex-vivo & Not Available \\
\Xhline{0.9pt}

\textbf{BREATH (ours)}
& Airway & 62 & 146{,}738 & In-vivo & Localization \\
\Xhline{1.2pt}
\end{tabular}
\end{table*}

To address the challenge in data scarcity, we first build BREATH dataset, the largest in-vivo endoscopic localization dataset, to the best of our knowledge, collected within the human airway during routine clinical procedures. It provides dense annotations including 3D models, depth, pose, and anatomy, as well as localization-oriented visual question answering (VQA) to support vision-language modeling for endoscopic localization and navigation.

Based on our dataset, we develop \textbf{BREATH-VL} (\textbf{B}ronchoscopy \textbf{REA}soning and \textbf{T}racking via \textbf{H}ybriding with \textbf{V}ision-\textbf{L}anguage models), a framework for robust 6-DoF bronchoscope localization. It combines semantic reasoning from a VLM, with vision-based geometric registration to achieve precise 6-DoF pose estimation. Specifically, the VLM provides coarse semantic localization by jointly detecting anatomical landmarks, estimating branch-level position and insertion depth, describing the scene in natural language.
The vision-based module integrate depth estimation and anatomical landmark detection to register the endoscope to a pre-reconstructed airway map and recover its 6-DoF pose in the CT coordinate. The overall design follows a dual-process reasoning principle. The VLM performs deliberate and context-aware reasoning, while the geometric modules carry out precise estimation. Through this complementary design, BREATH-VL achieves robust and accurate localization overcoming diverse visual degradation.

To further enhance the VLM’s semantic understanding with temporal information, we introduce a lightweight context-learning mechanism that encodes the endoscope’s recent motion history as linguistic prompts. This textual representation of temporal context allows the VLM to exploit motion cues and temporal correlations for more accurate localization, without the computational overhead of video-based inference. Consequently, BREATH-VL attains temporally consistent, anatomically aware semantic reasoning while maintaining efficient inference speed.

Meanwhile, we formally define the bronchoscopy scene estimation and localization task and introduce evaluation metrics to assess both coarse localization accuracy and full 6-DoF camera localization. Extensive experiments validate the effectiveness of BREATH-VL, demonstrating its strong semantic reasoning and localization capability in complex bronchoscopy scenes. Building on this foundation, BREATH-VL surpasses state-of-the-art vision-only bronchoscopy localization methods, achieving higher precision and robustness, and showing promising potential for integration into real clinical workflows.

The contributions of this work are as follows:

\begin{itemize}
    \item We formally define the bronchoscopy scene reasoning and localization task, and develop the largest bronchoscopy localization dataset and benchmark with comprehensive evaluation metrics for both coarse anatomical reasoning and fine-grained 6-DoF pose estimation.
    
    \item We propose BREATH-VL, a dual-loop localization framework that integrates semantic priors of vision-language model with vision-based methods, enabling both strong vision-language semantics and fine-grained localization.

    \item To further enhance the VLM’s semantic reasoning capability, we introduce a lightweight context-learning mechanism that encodes motion history as linguistic prompts, enabling efficient exploitation of temporal information. 
    
    \item Extensive experimental results demonstrate that BREATH-VL provides strong semantic reasoning in challenging surgical scenes, and that it outperforms state-of-the-art vision-only 6-DoF localization methods in both accuracy and robustness.
\end{itemize}
\section{Related Work}
\label{sec:relatedwork}

\subsection{Surgical Endoscopic Localization Dataset}
Computer-assisted endoscopic localization and navigation promise faster, more comprehensive examinations \cite{cold2024artificial} and support autonomous robotic operations \cite{banach2025conditional,sganga2019autonomous}. This potential has driven the release of several public datasets for endoscopic localization, as summarized in Table~\ref{tab:datasets}. However, most existing datasets are acquired under simulated environments \cite{bobrow2023,rau2024simcol3d,fulton2020comparing}, in phantoms \cite{golhar2025c3vdv2,deng2023feature}, or with ex-vivo specimens \cite{ozyoruk2021endoslam}, where the imaging domain differs substantially from real clinical scenes. Even for in-vivo datasets such as EndoMapper \cite{azagra2023endomapper}, pose annotations are available only for limited sequences.
Furthermore, most datasets focus on relatively simple anatomies such as the colon or stomach. Deng \textit{et al.} \cite{deng2023feature} introduced a bronchoscopy dataset with more complex airway structures for visual odometry, yet only phantom data are publicly available and no 3D models are provided for geometric-aware localization. 
In contrast, we present BREATH dataset, the largest in-vivo endoscopic localization dataset, to the best of our knowledge, collected within the complex human airway. It provides comprehensive annotations, including depth, pose, calibration, and 3D models, to support research on localization and reconstruction. Furthermore, we are the first to incorporate localization-oriented visual question answering (VQA), enabling vision-language modeling in surgical environments to assist localization and navigation.

\subsection{Vision-based Surgical Endoscopic Localization}
To realize the potential of computer assisted endoscopic localization, vision-based approaches have been developed, focusing on joint pose regression \cite{ozyoruk2021endoslam, sheikh2025endo, manni2024bodyslam, recasens2021,mackute2025navigational}, Gaussian splatting \cite{wu2025endoflow, wang2024endogslam, shao2022self}, registration \cite{mori2002tracking, deguchi2009selective, shen2019context, banach2021visually, sganga2019offsetnet,shu2025bronchopt}, retrieval \cite{zhao2019generative, sganga2019autonomous, banach2025conditional}, feature-based \cite{deng2023feature, borrego2023bronchopose}, and hybrid methods \cite{tian2024dd, tian2024pans, tian2025harnessing}. Despite their promising results, many of these methods are still limited to controlled, preclinical settings. They often struggle when faced with longer sequences or highly complex anatomies such as the human airway \cite{tian2024pans}. Visual challenges such as occlusions, anatomical deformation, and low-texture regions increase the difficulty of maintaining accurate localization over time. These limitations underscore the need for approaches capable of reasoning about complex scene contexts and integrating high-level anatomical semantics to ensure robust localization.

\subsection{Vision-Language Models for Localization and Navigation}
Vision-language models (VLMs) have demonstrated significant effectiveness in localization \cite{vivanco2023geoclip,xu2024addressclip,jia2025towards,cheng2025scale,xu2025addressvlm} and navigation \cite{an2024etpnav,tian2025drivevlm,yokoyama2025film} tasks, owing to their ability to integrate visual and semantic information. These models utilize semantic priors to enhance spatial predictions, grounding them in meaningful context that improves performance in complex environments. Notably, general-purpose VLMs \cite{bai2025qwen2,wang2024qwen2,chen2024internvl,liu2023llava,liu2024llavanext} have demonstrated significant efficacy through off-the-shelf or fine-tuning models to adapt to localization \cite{xu2025addressvlm} and navigation \cite{tian2025drivevlm,yokoyama2025film} tasks.

These developments indicate a promising direction for the application of VLMs in complex tasks, including endoscopic camera localization. By leveraging the semantic understanding capabilities of VLMs, it is possible to enhance the accuracy and robustness of localization systems in challenging environments. However, because VLMs are not designed for fine-grained continuous regression, existing work primarily uses them for address-level localization rather than precise pose estimation \cite{xu2024addressclip,xu2025addressvlm}. In our framework, we leverage VLMs’ semantic understanding for coarse camera localization, which guides geometric modules for 6-DoF pose regression, improving localization robustness and accuracy compared to vision-only methods.

\begin{table}[t]
\centering
\caption{Main Notations.}
\label{tab:notation}
\begin{tabular}{ll}
\toprule
Symbol & Description \\
\midrule
$\Omega$ & Airway mesh. \\
$T$ & Airway topological graph with anatomy labels. \\
$t$ & Time step. \\
$I_t$ & Endoscopic frame at time t. \\
\midrule
$s_t$ & 6-DoF bronchoscope pose at time $t$. \\
$s_t^0$ & Initial pose used to start registration at time $t$. \\
$s_t^*$ & Semantic pose proposal at time $t$. \\
\midrule
$B_t^k$ & The k-th detected anatomical branch. \\
$A_t$ & Predicted branch-level location at time $t$. \\
$p \in [0,1]$ & Normalized insertion depth along branch $A_t$. \\
\midrule
$L(\cdot)$ & Overall alignment cost. \\
$L_{\mathrm{depth}}(\cdot)$ & Depth similarity term. \\
$L_{\mathrm{lmk}}(\cdot)$ & Landmark alignment term. \\
$L_{\mathrm{ctr}}(\cdot)$ & Centerline constraint term. \\
$\alpha_1,\alpha_2,\alpha_3$ & Weights for $L_{\mathrm{depth}}$, $L_{\mathrm{lmk}}$, and $L_{\mathrm{ctr}}$. \\
\bottomrule
\end{tabular}
\end{table}

\begin{figure*}[tbp]
\centerline{\includegraphics[width=\textwidth]{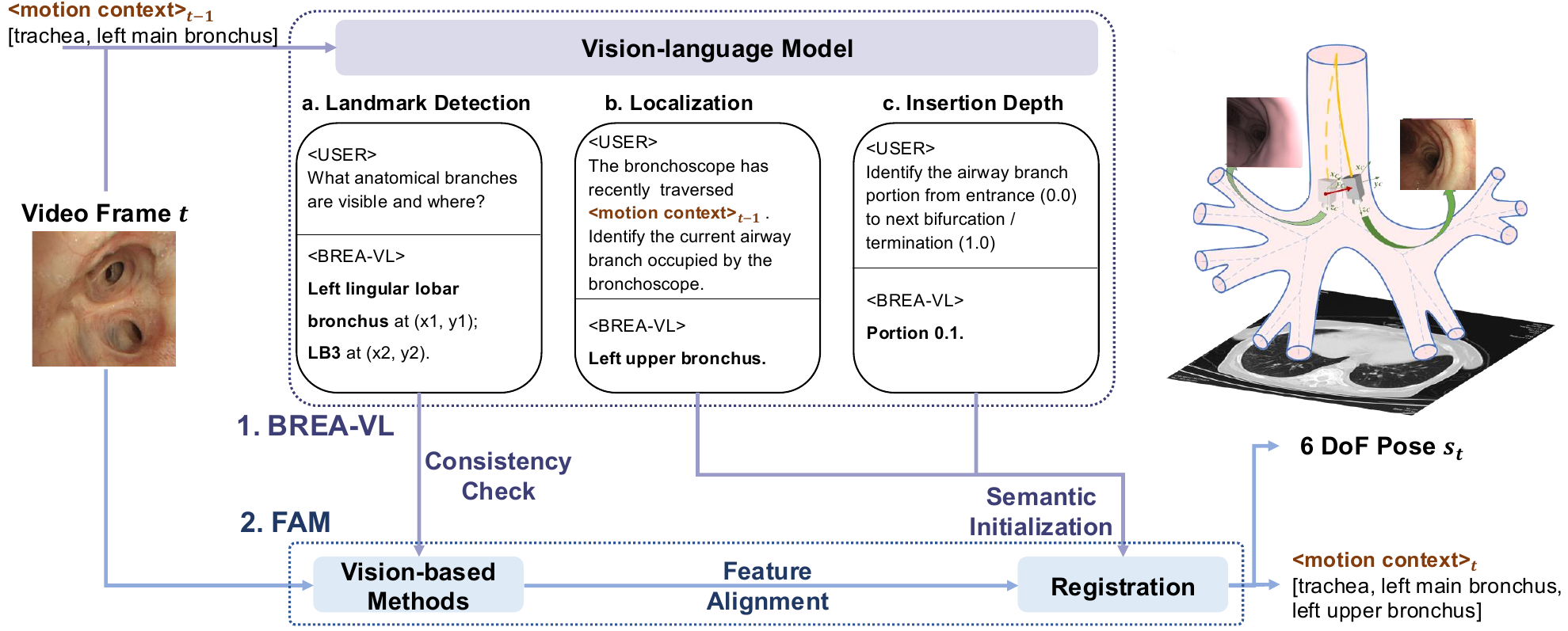}}
\caption{Overview of BREATH-VL for 6-DoF bronchoscopy localization. At time $t$, the bronchoscope pose $s_t$ is initialized with a semantic prior from BREA-VL and then refined via registration using vision-only geometric features.}

\label{fig:BREATH-vl}
\end{figure*}

\section{Problem Statement}
To facilitate reading, the main notations used in this work are presented in Table~\ref{tab:notation}.

In 6-DoF bronchoscopy localization, we operate on patient-specific CT scans and intra-operative endoscopic video. To effectively leverage the CT scan as an operative map, we adopt a practical setup in which the patient’s airway is pre-operatively segmented. From the CT volume, we reconstruct an airway surface mesh $\Omega$ and its topological graph $T$ with anatomical labels using existing methods such as ~\cite{yang2024progressive,wang2024accurate}. The airway mesh $\Omega$ provides geometric structure, while the graph $T$ supplies semantic cues that describe the topology of the operating space.  

During the intervention, we continuously receive an RGB observation $I_t$ from the endoscopic camera at each time step $t$. Our objective is to estimate the 6-DoF endoscopic camera pose $s_t$ at each time step as

\begin{equation}
    s_t = f\!\left(\{I_\tau\}_{\tau=1}^{t}, \Omega, T\right),
\end{equation}
where $f(\cdot)$ denotes a generic localization function that maps the observed endoscopic view and the patient-specific airway representation to a camera pose.

\section{Methods}
\label{sec:methods}

\subsection{Framework Overview}
BREATH-VL is a hybrid localization framework that combines semantic reasoning from a vision-language model with vision-based geometric registration to achieve accurate and robust 6-DoF camera pose estimation in bronchoscopy, as shown in Figure~\ref{fig:BREATH-vl}.

To ensure efficient and accurate pose estimation, BREATH-VL relies on two key components: 1) a semantic initializer, powered by BREA-VL (Section \ref{subsec:vl}), that predicts a coarse initial pose based on branch-level location and insertion depth with a contextual motion prompt for improved reliability. This module leverages language-guided motion context to overcome visual ambiguity and efficiently improve temporal continuity. Unlike traditional methods that warm-start from the previous frames, our approach avoids error accumulation and local minima by using the semantically informed initialization from BREA-VL. 2) A feature alignment module (FAM) (Section \ref{subsec:geo}), which refines the initial pose by registering the current endoscopic frame $I_t$ to the patient-specific CT representation. It leverages complementary visual cues, including depth and anatomical landmarks, to achieve accurate and reliable pose refinement.

This combination of high-level semantic inference and low-level geometric refinement allows BREATH-VL to maintain robustness in the presence of visual degradation, rapid camera movement, and anatomically repetitive structures. Crucially, the framework generalizes across patient cases without requiring per-case retraining, making it suitable for real-world surgical deployment.

\subsection{BREA-VL}
\label{subsec:vl}

BREA-VL is a vision-language model designed to perform bronchoscopy scene reasoning. It analyses through three complementary tasks: (1) anatomical landmark detection, (2) branch-level localization, and (3) insertion depth estimation. Then, the predictions are used to perform consistency check, and to generate a semantic initialization for later fine geometric optimization.

\textbf{Anatomical Landmark Detection.}
Landmarks in the airway, such as anatomical branch bifurcations, are crucial for coarse localization. Since the airway has a tree-like topology, the current camera position can be approximated by the visible branches. BREA-VL is prompted to describe the scene linguistically, including which anatomical structures are observed. Each detected branch $B_k$ is represented by a tuple $(a_k, x_k, y_k)$, where $a_k$ is the branch name and $(x_k, y_k)$ denotes its image coordinates.

\textbf{Branch-Level Localization with Contextual Prompting.}
Prior works have shown that branch-level localization is an effective way to narrow down the search space before fine pose refinement \cite{tian2024bronchotrack,tian2024pans}. They also demonstrate the importance of temporal context for disambiguating similar-looking regions \cite{tian2025endomamba}. While VLMs can take video clips as input to reason temporally, this introduces high computational cost and challenges in selecting meaningful keyframes under variable motion speeds.

To address this, we propose a simple but effective contextual prompting mechanism. Specifically, we record the anatomical branches traversed over time as a history sequence $A = {A_1, A_2, ..., A_{t-1}}$. We then construct a prompt describing this motion history and instruct BREA-VL to estimate the current branch $A_t$ based on both the scene and trajectory context. In practice, only the last three visited branches are included to preserve contextual relevance while minimizing token usage.

\textbf{Insertion Depth Estimation.}
Within a given branch, the endoscopic view varies significantly with insertion depth. To further localize the camera, BREA-VL is prompted to estimate the normalized depth $p \in [0, 1]$ along the predicted branch $A_t$. This provides a finer-grained position estimate and improves the accuracy of the semantic initializer for downstream optimization.

\textbf{Semantic Initialization.} 
Given the predicted branch-level location $A_t$ and the estimated insertion depth $p \in [0,1]$, we determine a semantic initialization point within the airway mesh. Specifically, we extract the centerline of branch $A_t$ and compute the 3D position along this path corresponding to the normalized depth $p$. Let this position be $(x_p, y_p, z_p)$. We then construct an initial pose estimate by combining this location with the most recent rotation estimate:

\begin{equation}
s_t^* = (x_p, y_p, z_p, r_{x}^{t-1}, r_{y}^{t-1}, r_{z}^{t-1}),
\end{equation}

\noindent where $(r_{x}^{t-1}, r_{y}^{t-1}, r_{z}^{t-1})$ are the roll, pitch, and yaw values from the optimized pose at time $t-1$.


Since BREA-VL operates at a lower frequency than the geometric optimizer, we only update the initialization with $s_t^*$ when a new semantic prediction is available. 


The final initial pose $s_t^0$ used for optimization is therefore defined as:


\begin{equation}
s_t^0 =
\begin{cases}
s_t^*, & \text{if } \delta_t = 1, \\
s_{t-1}, & \text{otherwise}.
\end{cases}
\label{eq:init}
\end{equation}

\noindent where $\delta_t \in {0, 1}$ denote an indicator variable, where $\delta_t = 1$ if BREA-VL has provided a valid update at time $t$, and $\delta_t = 0$ otherwise. This consistency check ensures BREA-VL provides reliable initial pose and enables more robust and accurate pose estimation.

Through the semantic initialization process, BREATH-VL improves optimization convergence and robustness, particularly in anatomically ambiguous or visually degraded regions, where tracking-based initialization often fails.

\begin{figure*}[tp]
\centerline{\includegraphics[width=\textwidth]{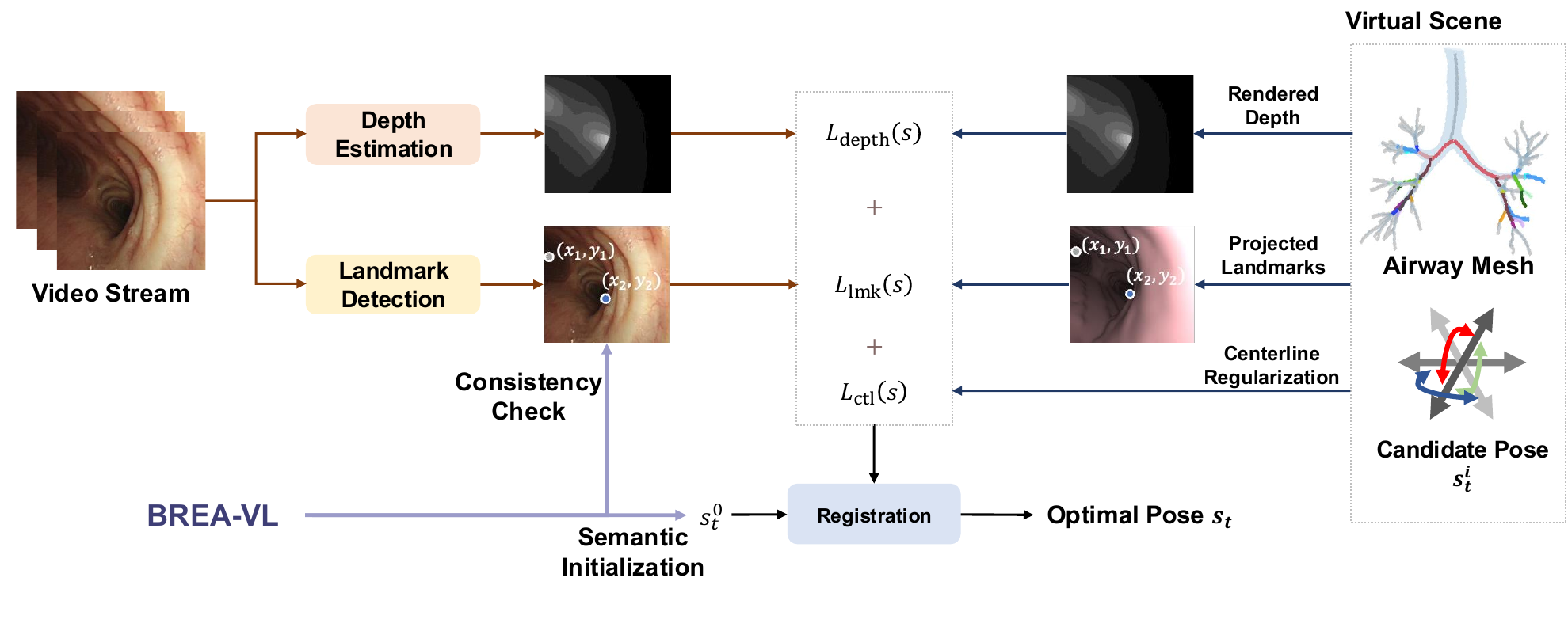}}
\caption{Overview of FAM for fine-grained bronchoscope localization. At time $t$, the bronchoscope pose $s_t$ is estimated by optimizing a composite objective that combines depth similarity, landmark alignment, and a centerline constraint. Localization accuracy is further improved through interaction with BREA-VL, which provides semantic pose initialization and landmark consistency checking.}
\label{fig_FAM}
\end{figure*}

\subsection{Feature Alignment}
\label{subsec:geo}
Figure \ref{fig_FAM} summarizes our feature alignment module (FAM). To obtain 6 DoF bronchoscope pose, FAM measures similarity between a pair of real and virtual bronchoscopy image by alignment cost $L(s)$. A candidate pose $s$ is scored by a weighted objective that combines three complementary cues: (i) depth-map agreement for geometry, (ii) landmark reprojection consistency for semantic disambiguation, and (iii) a centerline prior for physically plausible navigation:
\begin{equation}
    L(s) \;=\; \alpha_1\,L_{\mathrm{depth}}(s) \;+\; \alpha_2\,L_{\mathrm{lmk}}(s) \;+\; \alpha_3\,L_{\mathrm{ctr}}(s),
    \label{eq:pans_total}
\end{equation}
\noindent where $\alpha_1 = 0.5$, $\alpha_2 = 0.1$ and $\alpha_3 = 1.0$ are weights to balance the cost components.

The pose at time $t$ is obtained by minimizing the alignment cost by

\begin{equation}
s_t = \arg\min_{s} L(s).
\label{eq:opt}
\end{equation}

We use Powell’s derivative-free optimizer~\cite{fletcher1963rapidly} because the objective blends rendering, detection, and robust costs that are non-smooth and lack reliable gradients. The optimization is instructed by BREA-VL via providing the initial value $s_t^0$ by eq. \ref{eq:init} for improved robustness and accelerate convergence.

\textbf{Depth Similarity.}
To align geometry while remaining robust to illumination and texture changes, we compare the rendered depth from the virtual airway to depth inferred from the frame. We estimate per-frame depth with EndoOmni~\cite{tian2024endoomni}, a foundation model trained on large, diverse endoscopy data, which generalizes well across scopes and anatomies. Formally, we denote the depth estimation network as $G$ and compute the estimated depth as $z = G(I_t)$. 

Since the predicted depth is defined up to an unknown scale, we adopt normalized cross-correlation (NCC), which is scale- and bias-invariant, between the estimated depth $z$ and the depth rendered from the airway mesh $\Omega$ at camera pose $s$, denoted by $\bar z = Z(s, \Omega)$:
\begin{equation}
    L_{\mathrm{depth}}(s) \;=\; 1 - \mathrm{NCC}\!\big(z,\bar z\big),
\end{equation}
\noindent where $\mathrm{NCC}\!\big(z,\bar z\big)$ is the normalized cross-correlation between two depth maps, calculated with:
\begin{equation}
    \mathrm{NCC}\!\big(z,\bar z\big)=\; \frac{\sum_i (z_i-\mu_z)\,(\bar z_i-\mu_{\bar z})}
    {\sqrt{\sum_i (z_i-\mu_z)^2}\;\sqrt{\sum_i (\bar z_i-\mu_{\bar z})^2}},
\end{equation}
\noindent where $\mu_{z}, \mu_{\bar{z}}$ are the respective means. The term $L_{\mathrm{depth}}$ is small when the two depth maps are strongly correlated, indicating close geometric alignment.

\textbf{Landmark Alignment.}
Depth alone is often ambiguous in visually similar tubular regions and near bifurcations. To reduce this ambiguity, we first detect anatomical landmarks using EndoMamba~\cite{tian2025endomamba}, a video foundation model with a Mamba-based backbone that fuses spatial and temporal cues. Given image $I_t$ and hidden state $h_{t-1}$, the detector outputs landmark visibilities and image coordinates:
\begin{equation}
    f_{\mathrm{lmk}}^{\text{anat}}(I_t, h_{t-1}) = (\bar{\mathbf{M}}_t^{\text{anat}}, h_t),
\end{equation}
\begin{equation}
    \bar{\mathbf{M}}_t^{\text{anat}} = \big[(v_i, x_i, y_i)\big]_{i=1}^{n},
\end{equation}
where $v_i \in [0,1]$ is a visibility score for the $i$-th predefined anatomical branch and $(x_i, y_i)$ are its 2D image coordinates. 

We further enforce consistency with the anatomical landmarks predicted by BREA-VL. Let $c_i \in \{0,1\}$ be a consistency mask that is $1$ only if the $i$-th landmark agrees with the BREA-VL output, and define
\begin{equation}
    w_i = v_i c_i.
\end{equation}

We only retain landmarks with visibility probability greater than $0.5$, and get detection results:

\begin{equation}
    \mathcal{I} = \{\, i \mid w_i > 0.5 \,\},
    \quad
    \mathbf{M}_t^{\text{anat}} = \big[(a_i, x_i, y_i)\big]_{i \in \mathcal{I}},
\end{equation}

\noindent where $a_i$ is the anatomy-and hierarchy-aware branch label, and $(x_i, y_i)$ are the 2D coordinates.

To extend landmark coverage to distal peripheral branches without standard anatomical names, we additionally use a lumen tracker following BronchoTrack~\cite{tian2024bronchotrack}. By detecting lumens hierarchically, tracking them over time, and mapping them onto the patient-specific airway topology, we obtain branch labels and image locations:
\begin{equation}
    f_{\mathrm{lmk}}^{\text{lumen}}(I_t, \mathbf{M}_{t-1}^{\text{lumen}}, T) = \mathbf{M}_t^{\text{lumen}},
\end{equation}
\begin{equation}
    \mathbf{M}_t^{\text{lumen}} = \big[(a_j, x_j, y_j)\big]_{j=1}^{m},
\end{equation}
\noindent where $a_j$ is the hierarchy-aware branch label in the patient-specific airway tree, $(x_j, y_j)$ are the 2D coordinates of the corresponding lumen, $\mathbf{M}_{t-1}^{\text{lumen}}$ is the tracking results from the previous time step, and $T$ is the airway topology.

For a candidate pose $s$, we project the corresponding CT-defined 3D landmarks into the image as $(\hat x_i(s), \hat y_i(s))$ for anatomical landmarks and $(\hat x_j(s), \hat y_j(s))$ for distal lumens. The landmark alignment loss combines both sources:

\begin{equation}
\begin{aligned}
    L_{\mathrm{lmk}}(s)
    &=
    \frac{1}{\mathcal{I}}\sum_{i=1}^{\mathcal{I}} \big\|
        (x_i, y_i) - (\hat x_i(s), \hat y_i(s))
    \big\|_2 \\
    &+
    \frac{1}{m}
    \sum_{j=1}^{m}
    \big\|
        (x_j, y_j) - (\hat x_j(s), \hat y_j(s))
    \big\|_2,
\end{aligned}
\end{equation}

This encourages poses that are consistent with both semantically meaningful anatomical landmarks and distal lumen observations, improving robustness to false detections and generalization along deeper branches.

\begin{figure*}[tp]
\centerline{\includegraphics[width=\textwidth]{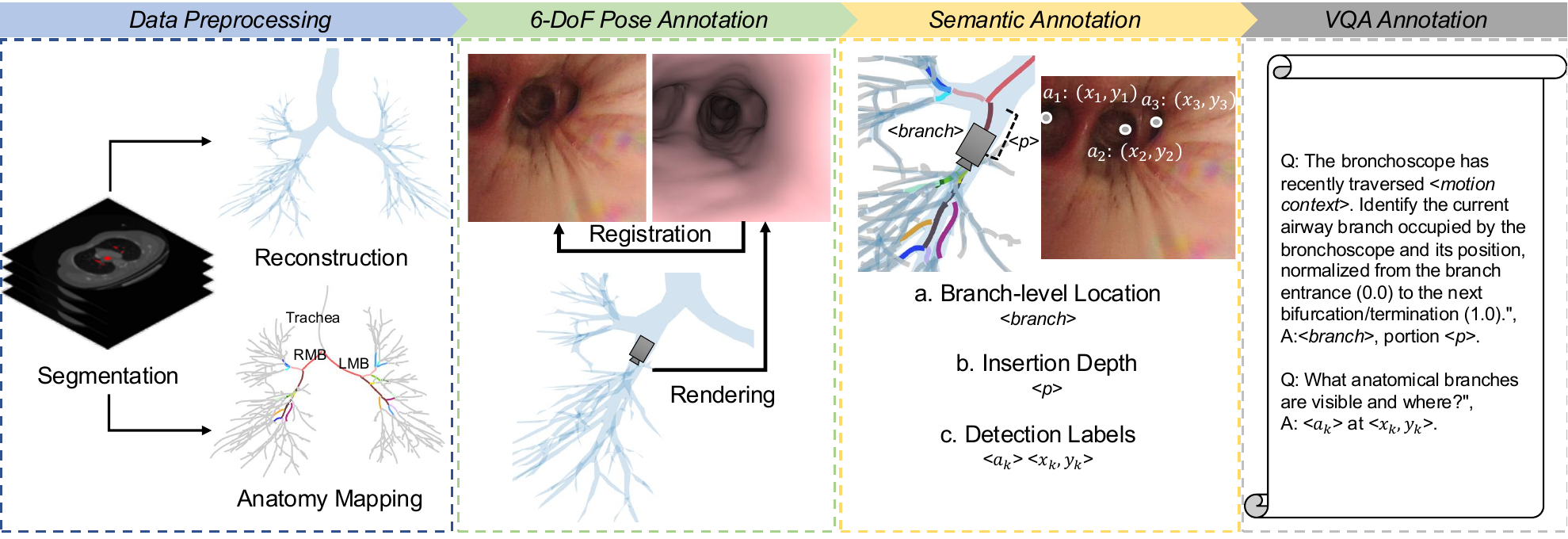}}
\caption{Data annotation pipeline for the BREATH dataset. After segmentation, 3D reconstruction, and anatomical mapping of the patient-specific airway, we first manually annotate the 6-DoF bronchoscope pose by registering virtual bronchoscopy images to real images. Using the labeled airway centerline, we then automatically generate semantic labels, including branch-level localization, insertion depth, and landmark detection. Finally, we convert these semantic labels into VQA annotations to build BREATH-VL.}
\label{fig:dataset_building}
\end{figure*}

\textbf{Centerline Constraint.}
Pose-only alignment can drift outside the lumen or to implausible viewpoints. We therefore impose a centerline prior to restrict the search to feasible trajectories. Let $d(s,A_t)$ be the shortest distance from the camera pose $s$ to the branch $A_t$ centerline, and $\phi(s,A_t)$ the angle between the optical axis and the local centerline tangent. We model both with zero-mean Gaussians and add their negative log-likelihoods:
\begin{equation}
    L_{\mathrm{ctr}}(s) = \mathcal{N}(d;\, 0, \sigma_1^2) \cdot \mathcal{N}(\phi;\, 0, \sigma_2^2),
\end{equation}
with $\sigma_1 = r/2$ and $\sigma_2 = \pi/6$, where $r$ is the radius of branch $b$. These settings encourage the scope to remain near the lumen center and roughly aligned with airway direction, while allowing natural maneuvering.


In this setup, the semantic module and the geometric module operate at two seperate threads: BREA-VL runs at a lower update rate, producing context-aware predictions intermittently, while the geometric module runs continuously to provide pose refinement.
This design effectively combines the strengths of both modules: BREA-VL contributes semantic grounding and robustness in ambiguous or visually degraded regions, while the geometric optimizer ensures frame-to-frame precision. Together, they form a complementary system that balances global context and local accuracy, enabling generalizable 6-DoF localization in challenging surgical scenes.
\section{BREATH Dataset}
\label{sec:dataset}
We present the BREATH dataset, the largest collection of calibrated recordings from routine bronchoscopies. BREATH contains 66 procedures performed on patients with various conditions such as pulmonary nodules, pneumonia, and lung cancer. Dataset contains patient-specific airway trimeshes reconstructed from preoperative CT, airway skeletons with anatomical topology, calibrated endoscope parameters, and per-frame 6-DoF endoscope pose labels in the CT coordinate system. All data were collected under IRB approval. In total, BREATH contains 148,926 pose-labeled frames. No prior public dataset captures real patient data from clinical workflow at this scale or with comparable annotations.

\subsection{Data Acquisition and Annotations}
For each procedure, we acquire three modalities: (1) a preoperative chest CT scan, (2) an in-procedure bronchoscopy video, and (3) checkerboard images for estimating intrinsic and distortion parameters. Bronchoscopy videos are recorded from the clinical endoscopy system at native resolution with frame rates between 10–20\,fps using four Olympus bronchoscopes. Each scope is calibrated from checkerboard images with the method of \cite{zhang1999flexible}.

The data annotation process is shown in Figure~\ref{fig:dataset_building}. First, CT volumes are used to reconstruct an airway surface mesh and to extract a centerline tree that preserves anatomical mapping, following \cite{yang2024progressive}. Then, we derive per-frame 6-DoF camera poses by aligning each bronchoscopy frame to the patient-specific CT geometry. To this end, we developed an OpenGL-based toolkit that loads the patient’s airway trimesh and instantiates a virtual camera whose intrinsics match the calibrated scope. Three trained annotators register the virtual views to the real images frame-by-frame, producing camera poses in the CT coordinate system. To assess annotation accuracy, two cases were independently labeled by all annotators, yielding a translational group variance of 0.58\,mm. 

Given the labeled poses and the airway skeleton, we assign each frame to the nearest airway branch to obtain branch-level localization and its insertion depth normalized to 0-1 in the corresponding branch. For visibility, we determine the set of branches expected to be in view and, for each visible branch, define its image-plane location by projecting the farthest visible centerline point. Finally, VQA labels are generated autonomously from the semantic annotations.

\subsection{Specifications}
Each case includes: a reconstructed airway mesh; one bronchoscopy video; camera calibration images; centerline graph with branch anatomy and hierarchy, and per-frame labels (6-DoF pose, branch-level location, visible-branch set). Across 66 procedures, 56 are used for training, and 10 are used for testing. All cases contain 148,926 pose-labeled frames. All poses are defined in the CT coordinate frame and are consistent with the provided intrinsics. 

\subsection{Tasks and Metrics}
BREATH supports three benchmark tasks with standardized evaluation protocols.

\begin{itemize}
    \item \textbf{Anatomical landmark detection.}
We report F1 scores for landmark detection. We regard a prediction accurate if its spatial error is within a threshold $\beta$.

Given a predicted landmark $\hat{\ell}_i \in \mathbb{R}^2$ and ground truth $\ell_i \in \mathbb{R}^2$ on the same frame, the Euclidean distance
\begin{equation}
    d_i = \lVert \hat{\ell}_i - \ell_i \rVert_2
\end{equation}
is required to satisfy
\begin{equation}
    d_i \le \beta \cdot \min(H, W),
\end{equation}
where $H$ and $W$ are the image height and width. After one-to-one matching, let $\text{TP}_\beta$, $\text{FP}_\beta$, and $\text{FN}_\beta$ denote the numbers of true positives, false positives, and false negatives. The F1 score at threshold $\beta$ is
\begin{equation}
    \mathrm{F1}_\beta = \frac{2\,\text{TP}_\beta}{2\,\text{TP}_\beta + \text{FP}_\beta + \text{FN}_\beta}.
\end{equation}

In our experiments, we report F1@0.1 and F1@1 by setting $\beta \in \{0.1, 1\}$. F1@0.1 emphasizes accurate spatial localization, whereas F1@1 primarily evaluates whether landmarks are correctly detected.

    \item \textbf{Branch-level localization.}
    For ground truth branch labels $a_i \in \{1,\dots,C\}$ and predictions $\hat{a}_i$, accuracy and macro-averaged F1 are used:
    \begin{equation}
        \mathrm{Acc}=\frac{1}{N}\sum_{i=1}^{N}\mathbf{1}[\hat{a}_i = a_i].
    \end{equation}
    Let $\text{TP}_c$, $\text{FP}_c$, and $\text{FN}_c$ be counts for class $c$. The per-class F1 is
    \begin{equation}
        \mathrm{F1}_c=\frac{2\,\text{TP}_c}{2\,\text{TP}_c+\text{FP}_c+\text{FN}_c},
    \end{equation}
    and the macro-F1 is
    \begin{equation}
        \mathrm{F1}=\frac{1}{C}\sum_{c=1}^{C}\mathrm{F1}_c.
    \end{equation}

    \item \textbf{Insertion depth.}
We report the mean absolute error (MAE) and root mean squared error (RMSE) of the predicted insertion depth, evaluated on correctly localized branches.

Let $p_t$ and $\hat{p}_t$ denote the ground-truth and predicted insertion depth at time $t$. 
The MAE and RMSE over correctly localized branches are defined as
\begin{equation}
    \mathrm{MAE} = \frac{1}{N} \sum \big| \hat{p}_t - p_t \big|,
\end{equation}
\begin{equation}
    \mathrm{RMSE} = \sqrt{ \frac{1}{N} \sum \big( \hat{p}_t - p_t \big)^2 }.
\end{equation}

    \item \textbf{6-DoF camera tracking.}
Given per-frame translations $T_t^{\mathrm{est}}, T_t^{\mathrm{gt}}$ and rotations $R_i^{\mathrm{est}}, R_i^{\mathrm{gt}}$ for a sequence of $N$ frames, we report the average translational Absolute Trajectory Error ($ATE_{trans}$):
\begin{equation}
    ATE_{trans}
    = \frac{1}{N}\sum_{i=1}^{N} \left\lVert T_t^{\mathrm{est}} - T_t^{\mathrm{gt}}\right\rVert_2.
\end{equation}
The average rotational Absolute Trajectory Error ($ATE_{rot}$) is computed as:
\begin{equation}
    ATE_{rot} = \frac{1}{N}\sum_{i=1}^{N} \arccos\!\left(\frac{\operatorname{tr}\!\left(R_i^{\mathrm{err}}\right)-1}{2}\right),
\end{equation}

\begin{equation}
    R_i^{\mathrm{err}} = \left(R_i^{\mathrm{gt}}\right)^{-1} R_i^{\mathrm{est}},
\end{equation}
where $i = 1,\dots,N$ indexes frames.

    \item \textbf{Tracking success rate.} 
    We report SR-5 and SR-10 as the fraction of frames with $ATE_{trans}$ below 5\,mm and 10\,mm respectively, following existing research \cite{shen2019context,banach2021visually,tian2024pans}:
    \begin{equation}
        \mathrm{SR-}\delta=\frac{1}{N}\sum_{t=1}^N \mathbf{1}\!\left[\left\lVert T_t^{\mathrm{est}}-T_t^{\mathrm{gt}}\right\rVert_2 \le \delta\right],
    \end{equation}
    
    \noindent where $\delta\in\{5,10\}\,\mathrm{mm}$.
\end{itemize}

\section{Experiments}
\label{sec:experiments}

\begin{table*}[tp]
  \centering
  \caption{6-DoF bronchoscope localization results on the BREATH dataset. Best performance for each metric is highlighted in \textbf{bold}.}
    \begin{tabular}{c|ccccccccccc}
    \toprule
    \multicolumn{1}{c|}{Trajectory} & Case1 & Case2 & Case3 & Case4 & Case5 & Case6 & Case7 & Case8 & Case9 & Case10 & Mean±Std \\
    \hline
    \multicolumn{1}{c|}{} & \multicolumn{11}{c}{$ATE_{trans}$ (mm) $\downarrow$} \\
    \hline
    EndoGSLAM \cite{wang2024endogslam} & 35.77 & 9.75  & 12.94  & 28.69  & 34.78  & 28.87  & 31.45  & 36.27  & 30.92  & 12.11  & 26.2±10.4 \\
  Endo-FASt3r \cite{sheikh2025endo} & 24.75 & 19.48  & 26.87  & 21.63  & 20.38  & 21.45  & 24.58  & 25.65  & 19.97  & 21.71  & 22.6±2.6 \\
  Depth-Reg \cite{shen2019context,banach2021visually} & 63.92 & 28.68  & 30.40  & 42.17  & 52.97  & 8.02  & 56.16  & 72.12  & 34.62  & 46.18  & 43.5±18.9 \\
   PANSv2 \cite{tian2025harnessing} & 8.89 & 10.11  & 10.47  & 9.80  & 10.04  & 10.55  & 11.15  & 13.29  & 9.35  & 8.08  & 10.2±1.4 \\
    BREATH-VL & \textbf{7.59} & \textbf{6.60} & \textbf{9.77} & \textbf{7.18} & \textbf{7.42} & \textbf{8.18} & \textbf{7.36} & \textbf{9.73} & \textbf{6.31} & \textbf{6.34} & \textbf{7.6±1.3} \\
    \hline
    \multicolumn{1}{c|}{} & \multicolumn{11}{c}{$ATE_{rot}$ (deg) $\downarrow$} \\
    \hline
    EndoGSLAM \cite{wang2024endogslam} & 101.70 & 38.06  & 63.57  & 76.85  & 96.60  & 77.90  & 75.62  & 81.58  & 70.13  & 53.86  & 73.6±18.8 \\
   Endo-FASt3r \cite{sheikh2025endo} & 76.67 & 78.49  & 83.25  & 79.11  & 76.82  & 81.80  & 79.81  & 78.80  & 78.24  & 78.48  & 79.1±2.0 \\
  Depth-Reg \cite{shen2019context,banach2021visually} & 98.66 & 123.61  & 110.27  & 137.89  & 101.10  & 82.08  & 134.46  & 131.50  & 59.08  & 113.59  & 109.2±25.0 \\
  PANSv2 \cite{tian2025harnessing} & \textbf{35.0} & 67.7  & 34.2  & 53.0  & 31.6  & 57.3  & 41.5  & \textbf{47.3} & \textbf{43.4} & \textbf{35.9} & 44.6±11.8 \\
   BREATH-VL & 35.1  & \textbf{67.2} & \textbf{30.1} & \textbf{32.7} & \textbf{29.6} & \textbf{45.3} & \textbf{37.0} & 63.9  & 55.1  & 37.2  & \textbf{43.3±14.0} \\
    \hline
    \multicolumn{1}{c|}{} & \multicolumn{11}{c}{SR-5 (\%) $\uparrow$} \\
    \hline
    EndoGSLAM \cite{wang2024endogslam} & 0.00 & 46.57 & 12.84 & 1.30 & 0.00 & 0.00 & 0.47 & 0.00 & 0.00 & 5.63 & 6.7±14.6 \\
   Endo-FASt3r \cite{sheikh2025endo} & 0.00 & 0.00 & 0.00 & 0.00 & 0.22 & 0.00 & 1.77 & 1.92 & 0.00 & 0.00 & 0.4±0.8 \\
   Depth-Reg \cite{shen2019context,banach2021visually} & 1.71 & 6.35 & 2.49 & 3.34 & 10.70 & 31.40 & 0.19 & 0.79 & 7.66 & 5.92 & 7.1±9.2 \\
   PANSv2 \cite{tian2025harnessing} & \textbf{46.32} & 46.72 & 41.54 & \textbf{48.55} & 38.86 & 34.08 & 34.74 & 21.54 & 29.53 & \textbf{51.75} & 39.4±9.5 \\
    BREATH-VL & 44.96 & \textbf{51.83} & \textbf{48.29} & 38.19 & \textbf{52.51} & \textbf{39.54} & \textbf{42.34} & \textbf{28.34} & \textbf{47.45} & 50.54 & \textbf{44.4±7.5} \\
    \hline
    \multicolumn{1}{c|}{} & \multicolumn{11}{c}{SR-10 (\%) $\uparrow$} \\
    \hline
    EndoGSLAM \cite{wang2024endogslam} & 0.9 & 68.4 & 45.5 & 9.6 & 0.8 & 3.8 & 0.8 & 0.9 & 0.1 & 61.1 & 19.2±27.7 \\
   Endo-FASt3r \cite{sheikh2025endo}& 0.0 & 0.0 & 0.0 & 3.3 & 5.3 & 0.0 & 4.4 & 4.9 & 0.0 & 0.0 & 1.8±2.4 \\
    Depth-Reg \cite{shen2019context,banach2021visually} & 3.8 & 14.0 & 17.0 & 5.6 & 21.3 & 60.9 & 3.8 & 1.5 & 16.0 & 12.5 & 15.6±17.2 \\
   PANSv2 \cite{tian2025harnessing} & \textbf{78.0} & 73.4 & 68.3 & 77.7 & 62.5 & 58.5 & 65.2 & 50.6 & 55.9 & 83.6 & 67.4±10.7 \\
   BREATH-VL & 76.3 & \textbf{80.7} & \textbf{70.0} & \textbf{80.5} & \textbf{76.8} & \textbf{66.4} & \textbf{76.8} & \textbf{62.2} & \textbf{83.9} & \textbf{84.7} & \textbf{75.8±7.4} \\
    \bottomrule
    \end{tabular}%
  \label{tab:result_6dof}%
\end{table*}%

\begin{figure*}[tbp]
\centerline{\includegraphics[width=\textwidth]{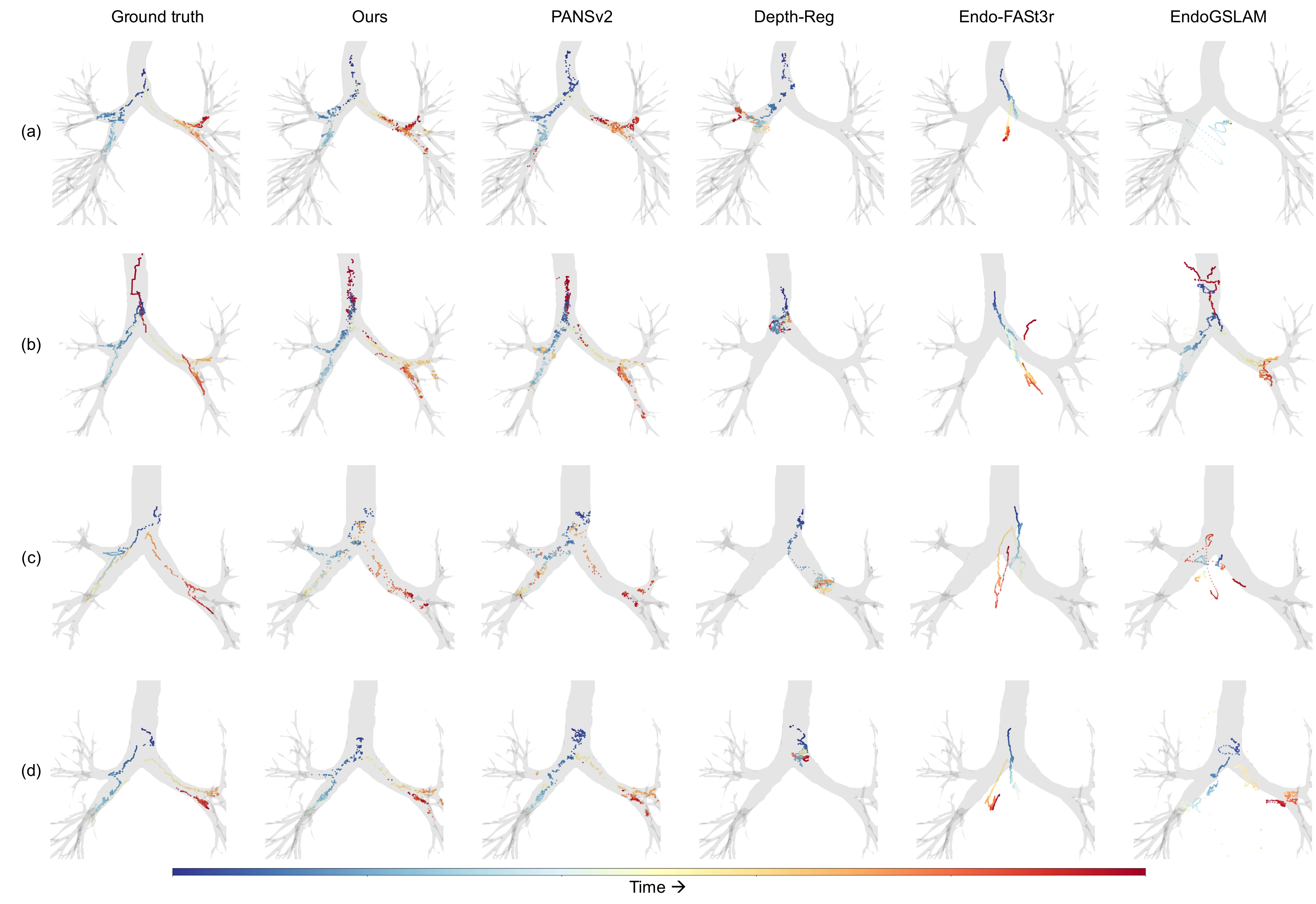}}
\caption{Localization trajectories overlaid on the airway mesh for four test patients, shown in subplots (a)-(d). A colormap encodes time along each trajectory, indicating the temporal order of camera poses and coverage of both sides of the airway. BREATH-VL produces trajectories that closely follow the ground truth, whereas PANSv2 often misidentifies landmarks and incurs large localization errors in deeper branches. Depth-Reg tracks the bronchoscope only over a short segment before failing due to depth ambiguity in similar anatomical regions, causing the optimization to become trapped in local minima. Endo-FASt3r gradually drifts away from the true camera pose because of its incremental localization strategy, while EndoGSLAM fails to reconstruct a complete scene under complex camera motion and the narrow field of view of the bronchoscope, resulting in large pose tracking errors.
}
\label{fig:traj}
\end{figure*}

\begin{figure*}[tp]
\centerline{\includegraphics[width=\textwidth]{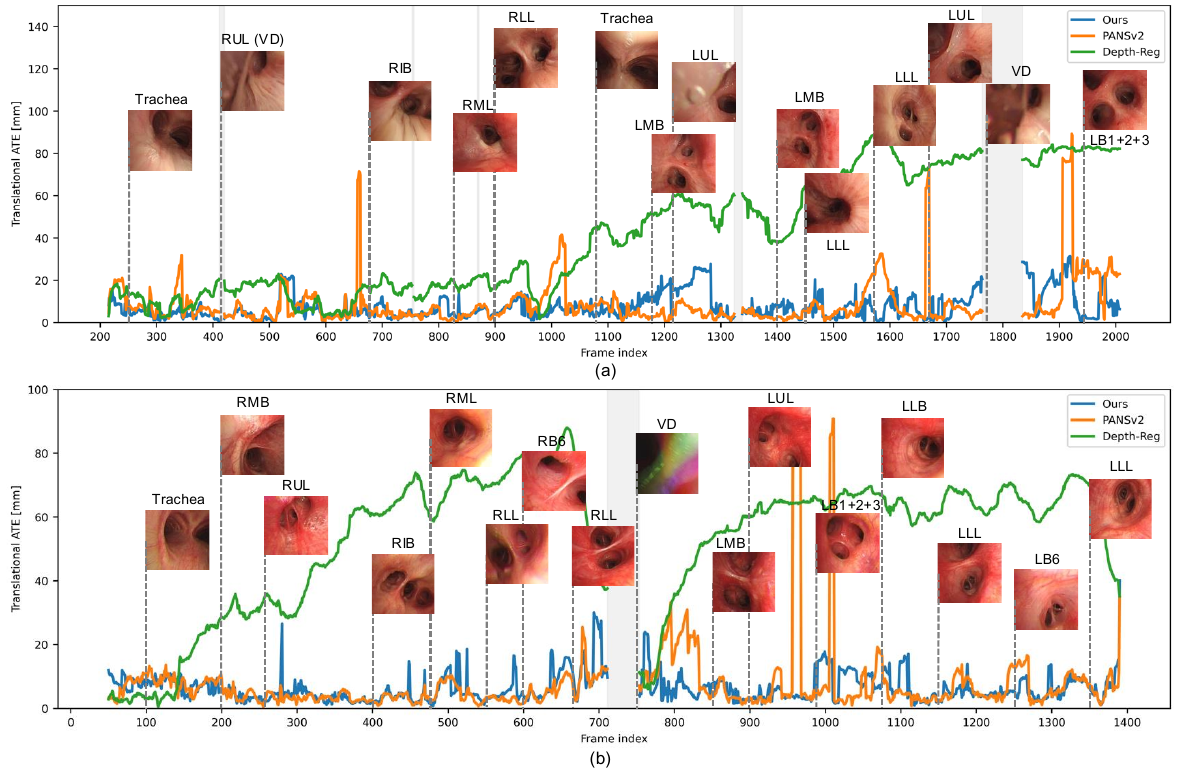}}
\caption{
Localization trajectory and $ATE_{trans}$ for example patient cases, shown in (a) and (b). Time steps without pose labels due to severe visual degradation are shown in light gray. Example bronchoscopic frames are shown at key anatomical landmarks: the trachea; right main bronchus (RMB), right upper lobe (RUL); right intermediate bronchus (RIB); right middle lobar bronchus (RML); RB6; right lower lobe (RLL); left main bronchus (LMB); left upper lobe (LUL); left lower lobe (LLL); left lingular bronchus (LLB); LB6; and the segment before LB1+2 and LB3 (LB1+2+3), as well as at frames affected by visual degradation (VD). BREATH-VL consistently maintains low translational error, whereas PANSv2 exhibits occasional large errors due to incorrect landmark recognition, and Depth-Reg tracks only short segments before becoming trapped in local minima.
}
\label{fig:error}
\end{figure*}

\begin{figure*}[tp]
\centerline{\includegraphics[width=\textwidth]{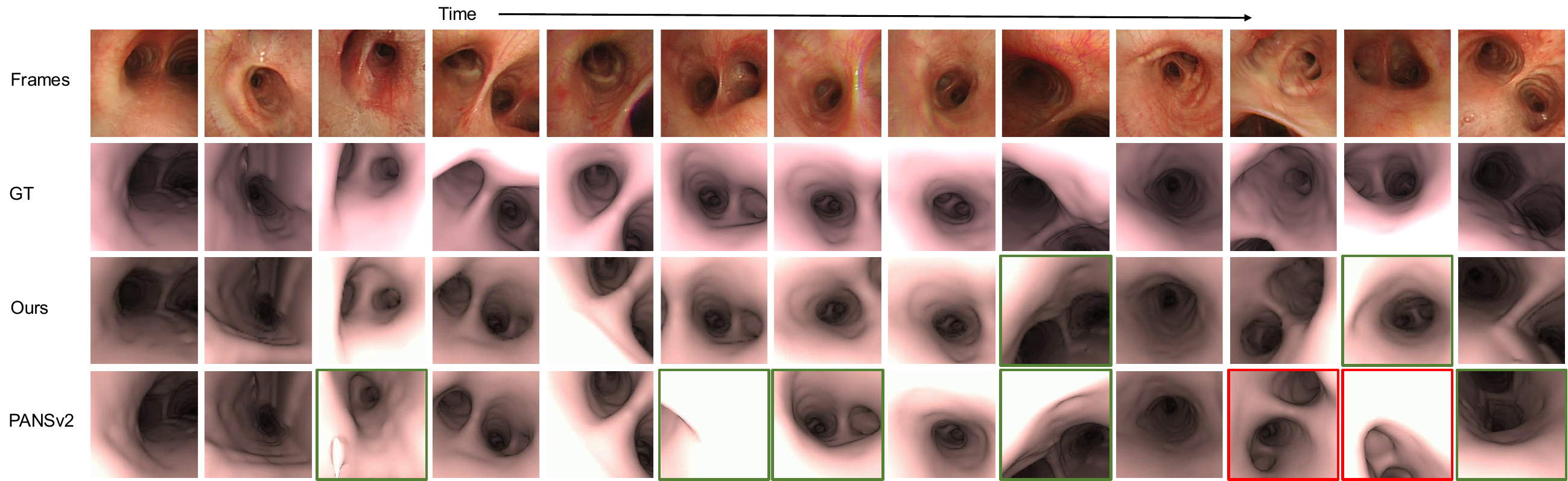}}
\caption{
Example virtual views localized by our BREATH-VL, comparing with best performing baseline PANSv2. Frames where the endoscope is assigned to the correct branch but exhibits large view misalignment are highlighted with \textcolor[rgb]{0,0.5,0}{green} boxes; frames where the endoscope is mislocalized to an incorrect branch and tracking is lost are highlighted with \textcolor{red}{red} boxes.
}
\label{fig:rendered_view}
\end{figure*}

\begin{figure}[tp]
\centerline{\includegraphics[width=\columnwidth]{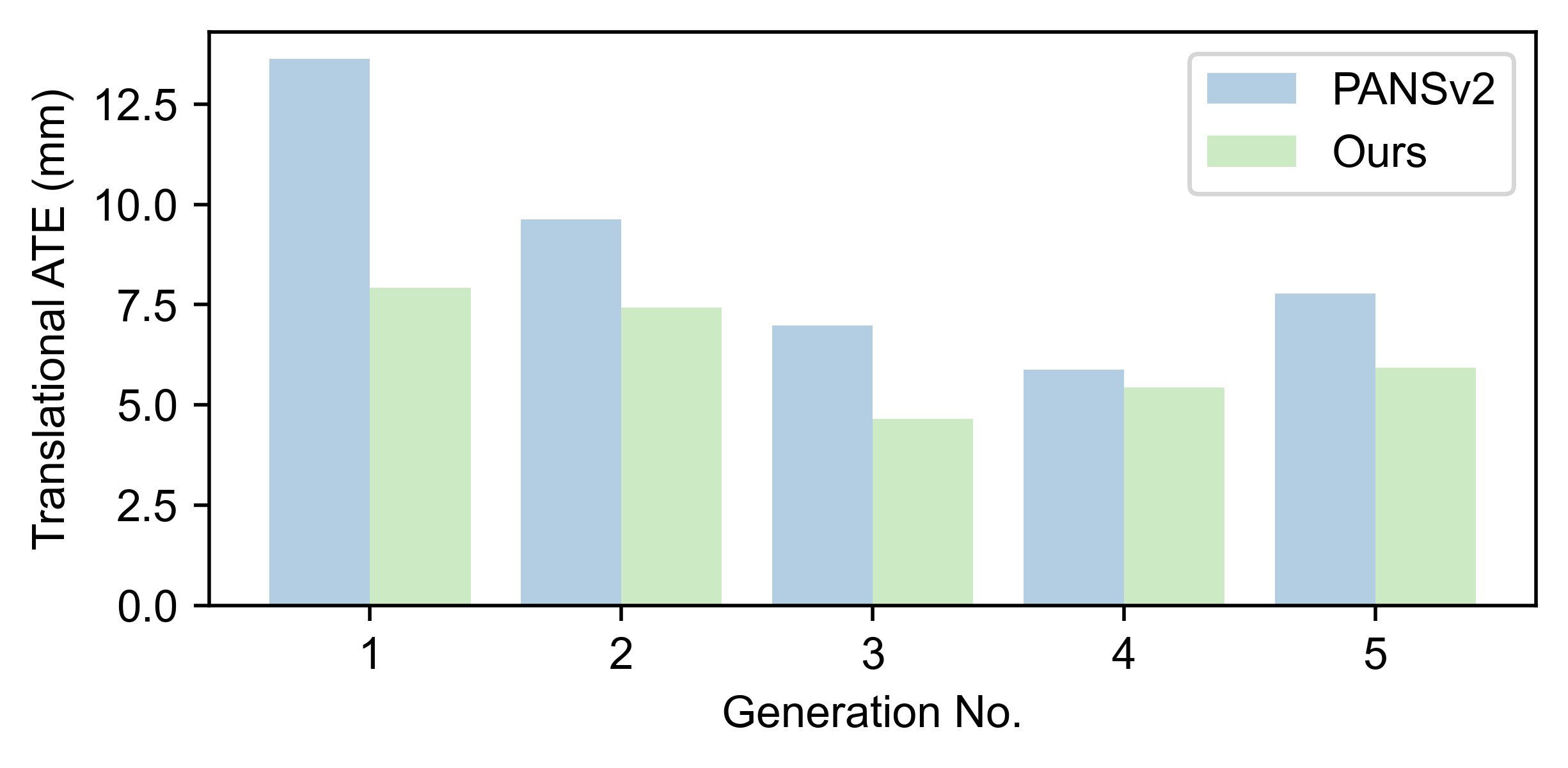}}
\caption{
Localization $ATE_{trans}$ across airway generations, comparing our BREATH-VL with the strongest baseline, PANSv2. BREATH-VL consistently achieves lower translational error than the vision-based PANSv2, across all airway generations.
}
\label{fig:branch_ate}
\end{figure}

In this section, we first describe the implementation details of BREATH-VL (Sec.~\ref{exp:implem}) and the baseline methods used for comparison (Sec.~\ref{exp:baselines}). We then evaluate 6-DoF localization accuracy against existing methods (Sec.~\ref{exp:6dof}). Next, we conduct ablation studies on the VLM backbone of BREA-VL, the motion-context prompt, and the use of video clips as VLM input (Sec.~\ref{exp:ablations}). Finally, we ablate the geometric registration module by comparing different alignment cost formulations and show that BREATH-VL consistently improves their performance by providing reliable initialization from BREA-VL (Sec.~\ref{exp:ablations}).

\subsection{Implementation Details}
\label{exp:implem}
We use InternVL3.5 \cite{wang2025internvl3} as the base vision-language model for BREA-VL. Pretrained on large-scale medical data, InternVL3.5 exhibits strong performance on surgical endoscopic data. To achieve faster inference, we adopt the 1.1B parameter variant, with 0.3B parameters in the vision encoder and 0.8B in the language model. We fine-tune BREA-VL using image frames resized to 448×448. During training, the endoscope motion context is generated from ground-truth endoscope poses. When integrating into the BREATH-VL localization system for inference, we instead use historically estimated endoscope poses to generate the motion context that guides BREA-VL.

\subsection{Baseline Methods}
\label{exp:baselines}
We compare the 6-DoF localization performance of BREATH-VL with existing surgical endoscopic pose estimation methods. Endo-FASt3r \cite{sheikh2025endo} is a self-supervised depth and pose estimation framework that leverages foundation models for endoscopic cameras. EndoGSLAM \cite{wang2024endogslam} localizes the endoscopic camera by reconstructing the surgical scene with Gaussian splatting \cite{kerbl20233d}. Depth-Reg \cite{shen2019context,banach2021visually} is a classic bronchoscopy localization method that optimizes camera pose through depth estimation and registration to the airway mesh. We re-implement Depth-Reg using EndoOmni \cite{tian2024endoomni} for endoscopic depth estimation and Powell's method \cite{fletcher1963rapidly} for registration to the airway mesh. PANSv2 \cite{tian2025harnessing} is a bronchoscopy localization framework that jointly optimizes the 6-DoF camera pose using depth estimation and landmark detection. PANSv2 is conceptually close to our FAM, but unlike PANSv2, we do not use any rule-based re-initialization module. 

For a fair comparison with learning-based methods such as Endo-FASt3r and PANSv2, we retrain their models on the BREATH dataset. Since EndoGSLAM requires RGB-D information, we additionally provide ground-truth depth as input. For scale-ambiguous methods, including Endo-FASt3r and EndoGSLAM, we align their predicted camera pose scale with the ground truth before evaluation. For bronchoscopy-specific methods, including PANSv2 and our BREATH-VL, we evaluate the results without any additional processing. Full inspection videos, from entering the trachea, through both sides of the peripheral airways, are used for testing without any manual frame filtering, making the experiment setup close to real clinical deployment.

\subsection{Results on 6-DoF Localization}
\label{exp:6dof}
Results on the 10 patient cases are reported in Table~\ref{tab:result_6dof}. BREATH-VL outperforms all competing methods across all metrics, achieving the lowest translation and rotation errors in every case and the highest tracking success rates. In contrast to geometry-aware methods, Endo-FASt3r and EndoGSLAM do not use the reconstructed airway map as input and therefore perform poorly on the BREATH dataset. Under rapid camera motion and large view-angle changes, the incremental tracking strategy of Endo-FASt3r gradually drifts away from the true camera pose, while EndoGSLAM fails to reconstruct a complete and globally consistent scene, resulting in large pose tracking errors. The corresponding estimated trajectories are shown in Figure~\ref{fig:traj}.

Among geometry-aware methods, BREATH-VL achieves the best overall accuracy and robustness. Figure~\ref{fig:traj} illustrates representative test-case trajectories against the airway meshes. Depth-based registration (Depth-Reg) is highly sensitive to local minima, often losing tracking in regions with weak geometric constraints or partial airway visibility. As a result, it only tracks the bronchoscope for a short segment. PANSv2 improves robustness through joint optimization with landmarks and leverages video input for landmark recognition, using temporal information to improve accuracy in challenging regions. However, due to the limited memory length of the video model, selecting informative keyframes that contain sufficient contextual information is challenging in real deployment, making landmark detection less reliable and causing performance degradation in long and complex examinations.
In contrast, BREATH-VL does not rely on explicit keyframe selection or separate landmark detectors. Instead, by using linguistic motion context as a prompt, it ensures that informative temporal cues are consistently provided to the model. By combining BREA-VL with the FAM, BREATH-VL continuously injects vision-language priors as global constraints into pose optimization, enabling stable tracking over long sequences. This design reduces sensitivity to local minima, mitigates drift accumulation, and maintains reliable localization even under rapid camera motion and large viewpoint changes. Because BREATH-VL provides a strong translational initialization for registration, the improvement is particularly pronounced in translational ATE. Figure~\ref{fig:error} shows two examples of translational ATE over time for representative cases. We also visualize virtual views localized by BREATH-VL and by the best-performing SOTA baseline, PANSv2. As shown in Figure~\ref{fig:rendered_view}, virtual views rendered from BREATH-VL poses align more closely with real endoscopic frames, with more accurate branch-level localization and supporting more precise downstream 6-DoF bronchoscopy localization. In addition, Fig.~\ref{fig:branch_ate} reports the translational error across airway generations, comparing BREATH-VL with PANSv2. BREATH-VL consistently reduces translational error at all generations, narrowing the search space in proximal, thicker branches and providing accurate branch recognition that improves localization in deeper, distal generations.

\begin{table*}[tbp]
  \centering
  \caption{Coarse localization and landmark detection results on the BREATH dataset. Best performance for each localization and detection metric is highlighted in \textbf{bold}. Insertion-depth error is reported only for correctly localized samples.}
    \begin{tabular}{c|c|cccc|cc|cc}
    \toprule
    \multirow{2}[2]{*}{Method} & \multirow{2}[2]{*}{Param.} & \multicolumn{4}{c|}{Branch-level localization} & \multicolumn{2}{c|}{Insertion Depth} & \multicolumn{2}{c}{Detection} \\
          &       & Precision$\uparrow$ & Recall$\uparrow$ & F1$\uparrow$    & Acc$\uparrow$ & MAE$\downarrow$   & RMSE$\downarrow$ & F1@1$\uparrow$  & F1@0.1$\uparrow$ \\
    \hline
    Finetuning MiniCPM-V-2 & 3B    & 0.130  & 0.112  & 0.111  & 0.520  & 0.284  & 0.349  & 0.348  & 0.040  \\
    Finetuning QwenVL3 & 2B    & 0.628  & 0.590  & 0.602  & 0.833  & 0.233  & 0.339  & 0.597  & 0.396  \\
    Finetuning InternVL3 & 1B    & \textbf{0.705} & 0.555  & 0.576  & 0.841  & 0.204  & 0.317  & 0.692  & 0.469  \\
    BREA-VL w/ InternVL3.5 &    1B   & 0.695  & \textbf{0.677} & \textbf{0.682} & \textbf{0.893} & \textbf{0.129} & \textbf{0.208} & \textbf{0.771} & \textbf{0.557} \\
    \bottomrule
    \end{tabular}%
  \label{tab:vlms}%
\end{table*}%

\begin{figure*}[tp]
\centerline{\includegraphics[width=\textwidth]{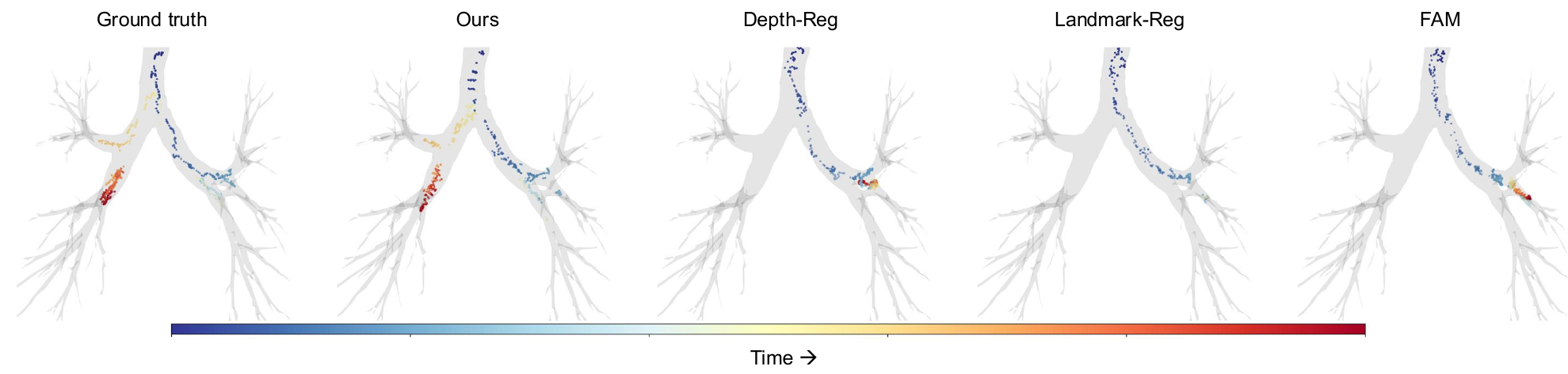}}
\caption{
Localization trajectories for a representative case, comparing BREATH-VL with registration-based methods. Purely registration-based approaches, using depth, landmarks, or mixed representations such as FAM, exhibit large errors over long trajectories, often becoming trapped in local minima and losing track under complex endoscope motion, whereas BREATH-VL maintains accurate tracking.
}
\label{fig:geo}
\end{figure*}

\begin{table}[tp]
\setlength{\tabcolsep}{4pt}
  \centering
  \caption{Ablations on Motion Context Prompt.}
    \begin{tabular}{c|cc|cccc}
    \toprule
          & \multicolumn{2}{c|}{Modules} & \multicolumn{4}{c}{Branch-level localization} \\
          & MC    & Seq   & Precision$\uparrow$ & Recall$\uparrow$ & F1$\uparrow$   & Acc$\uparrow$ \\
    \hline
    BREA-VL w/o MC &       &       & 0.575  & 0.578  & 0.571  & 0.795  \\
    BREA-VL w/ Seq &       & $\checkmark$ & 0.530  & 0.446  & 0.457  & 0.775  \\
    BREA-VL & $\checkmark$ &       & \textbf{0.695} & \textbf{0.677} & \textbf{0.682} & \textbf{0.893} \\
    \bottomrule
    \end{tabular}%
  \label{tab:motion_context}%
\end{table}%

\begin{table*}[tp]
  \centering
  \caption{Ablations on vision-based geometric methods in BREATH-VL. Best performance is highlighted in \textbf{bold}.}
    \begin{tabular}{ccccc}
    \toprule
    {Method} & $ATE_{trans}$ (mm) $\downarrow$ & $ATE_{rot}$ (deg) $\downarrow$  & SR-5 (\%) $\uparrow$  & SR-10 (\%) $\uparrow$ \\
    \midrule
    Depth-Reg & 43.5±18.9 & 109.2±25.0 & 3.9±3.6 & 13.5±17.8 \\
    BREATH-VL w/ Depth-Reg & 14.7±13.7 & 77.8±34.1 & 32.3±16.1 & 59.5±24.2 \\
    Landmark-Reg & 50.8±24.3 & 95.3±14.3 & 8.3±5.5 & 22.2±11.1 \\
    BREATH-VL w/ Landmark-Reg & 7.9±1.6 & 57.2±16.0 & 42.6±6.4 & 75.7±7.1 \\
    FAM   & 47.8±14.6 & 80.0±14.3 & 11.8±9.1 & 21.4±12.1 \\
    BREATH-VL w/ FAM & \textbf{7.6±1.3} & \textbf{43.3±14.0} & \textbf{44.4±7.5} & \textbf{75.8±7.4} \\
    \bottomrule
    \end{tabular}%
  \label{tab:ablations_on_geo}%
\end{table*}%

\begin{table}[tp]
  \centering
  \caption{Execution Time Statistics.}
    \begin{tabular}{cc|cc}
    \toprule
    Thread 1 & Times(ms) & Thread 2 & Times(ms) \\
    \hline
    Depth Estimation & 20     & \multirow{3}[2]{*}{BREA-VL} & \multirow{3}[2]{*}{240} \\
    Landmark Detection & 61    &       &  \\
    Registration & 92   &       &  \\
    \bottomrule
    \end{tabular}%
  \label{tab:speed}%
\end{table}%

\subsection{Ablation Studies}
\label{exp:ablations}
\textbf{Base Model of BREA-VL.}
We compare different vision-language base models for fine-tuning BREA-VL by replacing its backbone with several widely used architectures. Qwen3-VL \cite{yang2025qwen3} is a recent multimodal model family that extends the Qwen language backbone to vision inputs and supports strong general-purpose vision-language understanding and reasoning. MiniCPM-V-2 \cite{yao2024minicpm} is a lightweight vision-language model designed for efficient deployment, which balances recognition performance with low memory and computational cost. InternVL3 \cite{chen2024internvl} and InternVL3.5 \cite{wang2025internvl3} are two generations of high-performance vision-language models that integrate a strong visual encoder with a large language backbone. To accommodate the limited computational resources of surgical navigation systems and to improve inference speed, we adopt small variants of these models with fewer than 3B parameters. Results are reported in Table~\ref{tab:vlms}. Our BREA-VL, built on InternVL3.5, outperforms the variants based on Qwen-VL-3 and MiniCPM-V-2 in coarse localization. We attribute this advantage in part to additional pretraining of the InternVL family on medical data, which better aligns the model with endoscopic imagery. This improved coarse localization enables more accurate 6-DoF pose estimation in the subsequent registration stage.

\textbf{Motion Context Prompt.}
To demonstrate the effectiveness of using motion context as a text prompt to guide coarse localization, we conduct ablation studies in which we remove the motion context from the prompt. We also evaluate an alternative design that injects temporal information through vision by feeding short video clips. Specifically, we provide 4 frames sampled with a stride of 10 time steps as input to BREA-VL, without any linguistic motion context. Results are reported in Table~\ref{tab:motion_context}, where “w/o MC” denotes BREA-VL without motion-context prompt, and “w/ Seq” denotes without motion-context prompt and with frame-sequence input. Our linguistic motion context prompt significantly improves coarse localization performance, yielding much higher branch recognition F1 score and insertion depth accuracy. In contrast, using a short video clip does not consistently improve over single-frame input and still underperforms our motion-context design. We hypothesize that selecting keyframes that carry the most informative temporal cues is itself challenging, and simply feeding more frames may not add meaningful semantic information while making optimization harder. By explicitly encoding temporal semantics into a compact linguistic representation, the motion context enables BREA-VL to perform more accurate and robust coarse localization.

\textbf{Vision-only Geometric Methods.}
We use coarse localization from BREA-VL to guide several 6-DoF localization methods. In addition to FAM, BREA-VL provides initialization for depth-based registration \cite{shen2019context,banach2021visually} and landmark-based registration. As shown in Table~\ref{tab:ablations_on_geo}, registration-only methods exhibit large errors. A representative example in Figure~\ref{fig:geo} shows that these errors arise because each frame is optimized from the previous frame’s estimated pose: once the optimizer converges to an incorrect local minimum, especially under complex bronchoscopic motion or visual degradation, the error propagates forward and the tracker fails to recover.

Augmenting these registration methods with BREATH-VL markedly improves performance. The coarse pose from BREA-VL provides a semantically informed, temporally consistent initialization that reduces dependence on the previous frame and steers optimization toward the correct basin of attraction, leading to more robust tracking and fewer failures under rapid camera motion. Moreover, using a richer visual representation such as our FAM module further improves performance over single-representation baselines. This highlights the generality of our BREATH-VL framework: BREA-VL acts as a general enhancement layer for vision-only registration methods, with the potential to further benefit future, more advanced geometric pipelines.

\section{Discussion}
\label{sec:discussion}
Vision-based bronchoscopy localization faces significant challenges due to visual artifacts and the highly repetitive airway anatomy. Although prior works leverage various visual cues, such as depth, landmarks and visual odometry, their robustness under complex visual conditions remains limited. Consequently, they are typically evaluated only on manually curated sequences or controlled experimental data.
In this work, we propose BREATH-VL, a vision-language-guided 6-DoF bronchoscopy localization framework that robustly and accurately tracks the bronchoscope on full, clinically acquired sequences. We first leverage the strong semantic understanding of a vision-language model to obtain a coarse localization of the bronchoscope. To further improve performance and mitigate ambiguities caused by the repetitive airway anatomy, we encode temporal information as a motion-context prompt to the language model. We then apply a vision-only method that formulates bronchoscope localization as view-alignment registration between the bronchoscopic image and a preoperatively constructed CT-based map, yielding a precise 6-DoF pose. The high-level semantics provided by the vision-language model enable BREATH-VL to remain robust under visual degradation and to quickly recover from tracking failures once the view becomes clear. The low-level geometric registration of the vision-only method ensures precise localization by using the robust rough initialization from the vision-language model. By running the vision-language and vision-only modules synchronously, BREATH-VL achieves a favorable balance among robustness, accuracy, and computational efficiency.

Previous vision-only methods have explored leveraging temporal information for more accurate tracking. These include landmark-based approaches \cite{tian2024bronchotrack,tian2025harnessing}, which exploit lumen tracking across video frames to improve landmark recognition and mitigate ambiguity caused by similar anatomy, and pose-regression-based methods \cite{tian2024pans,tian2024dd,mackute2025navigational,deng2023feature}, which estimate camera motion between frames for incremental pose estimation or faster registration convergence.
However, due to limited computational resources and information decay over long sequences, these methods typically incorporate temporal information by selectively choosing a subset of frames for evaluation. In bronchoscopy, where the bronchoscope motion is highly irregular over the course of an intervention and visual artifacts frequently contaminate the field of view, selecting informative frames that provide effective temporal cues becomes a challenging problem in itself. This difficulty is closely related to the keyframe selection problem in SLAM \cite{younes2017keyframe} and recent work on video understanding \cite{tang2025adaptive}, where carefully choosing keyframes is crucial for strong performance. As a result, existing vision-based methods struggle to robustly localize the bronchoscope over long bronchoscopic videos.

Instead of focusing on keyframe selection, we propose a simple yet effective motion-context prompt for our vision-language model. By encoding the motion history into a linguistic motion context, information from long video sequences is naturally compressed into a textual description of the traversed trajectory. Our ablation study shows that this motion-context prompt substantially improves rough bronchoscopy localization performance of vision-language models, yielding lower trajectory error and reduced standard deviation across patient cases, indicating improved robustness.

Despite its superior accuracy and robustness, BREATH-VL still has several limitations. First, its localization speed is constrained. On a workstation with an NVIDIA GeForce RTX 4090 GPU and an Intel Core i9-14900 CPU, the system achieves an average runtime of approximately 5.6 frames per second (FPS). The execution time statistics are shown in Table \ref{tab:speed}. Although the vision-language and vision-only modules operate synchronously, the main bottleneck lies in refining the precise 6-DoF bronchoscope pose from the rough BREATH-VL initialization, which requires frequent depth rendering of candidate poses in the virtual environment. Second, integrating a language model into the localization pipeline incurs substantially higher memory usage compared to vision-only methods, with BREATH-VL requiring around 20 GB of GPU memory for inference. These limitations could be mitigated by adopting faster rendering and optimization strategies, as well as leveraging future hardware improvements.

\section{Conclusion}
In this work, we investigate the use of vision-language models (VLMs) for accurate and robust bronchoscopy localization. We first address data scarcity by constructing the BREA dataset, the largest in-vivo endoscopic localization dataset collected in the human airway during routine clinical procedures.
Building on this dataset, we propose BREATH-VL, a hybrid framework that combines the strong semantic understanding of a VLM for coarse localization with vision-based geometric registration for precise 6-DoF pose estimation. In this design, the VLM provides generalizable semantic cues that improve cross-patient adaptation and robustness against visual artefacts, while the vision-based registration refines these predictions to obtain accurate poses.
To further enhance accuracy and robustness by exploiting temporal information, we introduce a motion-context prompt that encodes the endoscope trajectory as a linguistic description, enabling efficient temporal reasoning without expensive video processing or complex keyframe selection.
Extensive experiments on complete, clinically collected surgical videos demonstrate that BREATH-VL achieves accurate and robust bronchoscope localization across diverse patient cases.




%


\bibliographystyle{IEEEtran}
\bibliography{main}

\newpage

 




\vfill

\end{document}